\documentclass[journal]{IEEEtran}
\usepackage{graphicx}
\usepackage{subfigure}
\usepackage{multirow}
\usepackage{longtable}
\usepackage{algorithm}
\usepackage{algorithmic}
\usepackage{booktabs}
\usepackage{stfloats}
\usepackage{caption}
\usepackage{amssymb,amsmath}
\usepackage{color}
\usepackage{fancyheadings}
\usepackage{hyperref}
\usepackage{cite}

\begin{document}
\title{Exposing Deepfake Face Forgeries with Guided Residuals}

\author{Zhiqing~Guo$^{1}$  \quad Gaobo~Yang$^{1}$ \quad Jiyou~Chen$^{1}$ \quad Xingming~Sun$^{2}$ \qquad \vspace{1pt}\\
$^{1}$Hunan University  \qquad $^{2}$Nanjing University of Information Science and Technology\qquad\qquad\\
\hspace{0.1in}{\tt\small \{guozhiqing, yanggaobo\}@hnu.edu.cn} \qquad {\tt\small cjyhn0302@gmail.com} \qquad  {\tt\small sunnudt@163.com} \\
}

\markboth{}%
{Shell \MakeLowercase{\textit{et al.}}: Exposing Deepfake Face Forgeries with Guided Residuals}

\maketitle

\renewcommand{\headrulewidth}{0pt}

\begin{abstract}
Residual-domain feature is very useful for Deepfake detection because it suppresses irrelevant content features and preserves key manipulation traces. However, inappropriate residual prediction will bring side effects on detection accuracy. In addition, residual-domain features are easily affected by image operations such as compression. Most existing works exploit either spatial-domain features or residual-domain features, while neglecting that two types of features are mutually correlated.
In this paper, we propose a guided residuals network, namely GRnet, which fuses spatial-domain and residual-domain features in a mutually reinforcing way, to expose face images generated by Deepfake.
Different from existing prediction based residual extraction methods, we propose a manipulation trace extractor (MTE) to directly remove the content features and preserve manipulation traces. MTE is a fine-grained method that can avoid the potential bias caused by inappropriate prediction. Moreover, an attention fusion mechanism (AFM) is designed to selectively emphasize feature channel maps and adaptively allocate the weights for two streams.
The experimental results show that the proposed GRnet achieves better performances than the state-of-the-art works on four public fake face datasets including HFF, FaceForensics++, DFDC and Celeb-DF. Especially, GRnet achieves an average accuracy of 97.72\% on the HFF dataset, which is at least 5.25\% higher than the existing works.
\end{abstract}
\IEEEpeerreviewmaketitle

\section{INTRODUCTION}
\IEEEPARstart{w}{ith} the rapid development of computer graphics (CG) and image processing, we should not take the credibility of visual content for granted any more \cite{CGR_tmm}. Especially, the recent progress of Deepfake has enabled fake face images to be scarily real.
Meanwhile, face images, which contain rich personal identity information such as gender, race, age and emotion, have been widely-used as the biological modality for access control such as the entrance to restricted area. Yet, the public availability of Deepfakes easily leads to misinformation. For example, online dating sites might use Deepfake-rendered faces to attract users or even cheat them \footnote{https://www.datingadvice.com/online-dating/generated-photos-can-make-dating-sites-more-welcoming}. Moreover, spies may use Deepfake-generated photos to create hard-to-detect fake profiles on LinkedIn \footnote{https://www.entrepreneur.com/article/335293}.

In recent years, many works have been presented to expose Deepfake face images. Among them, some works exploited hand-crafted features to represent biological signal inconsistencies, which include eye blinking \cite{eye_blinking}, head pose \cite{head_poses} and imprecise geometry shapes such as weird ears and asymmetrical faces \cite{visual_artifacts}. The other works introduced deep learning into face image forensics, which learn discriminative features from manipulation traces or biological inconsistencies. However, most of the existing deep learning based works exposed fake face images by directly learning spatial-domain features from RGB images \cite{MesoNet,AppliedSciences}, and the other works learned residual-domain features from prediction residuals to improve detection accuracy \cite{binary_highpassfilter,AMTEN}. In essence, learning features from prediction residuals is an effective way to highlight manipulation traces and suppress image content \cite{forensics1_tmm,forensics2_tmm}. Note that the residuals are obtained by either the fixed predictor \cite{pip1popescu,pip2qiu,pip3kirchner} or the learning-based predictor \cite{AMTEN}. Because there usually exists prediction bias for the existing predictors, the residual features are coarse-grained. Apparently, if the residuals can be directly extracted instead of prediction, it will be an intuitive and fine-grained way.

\begin{figure}
  \centering
  \includegraphics[width=3.4in]{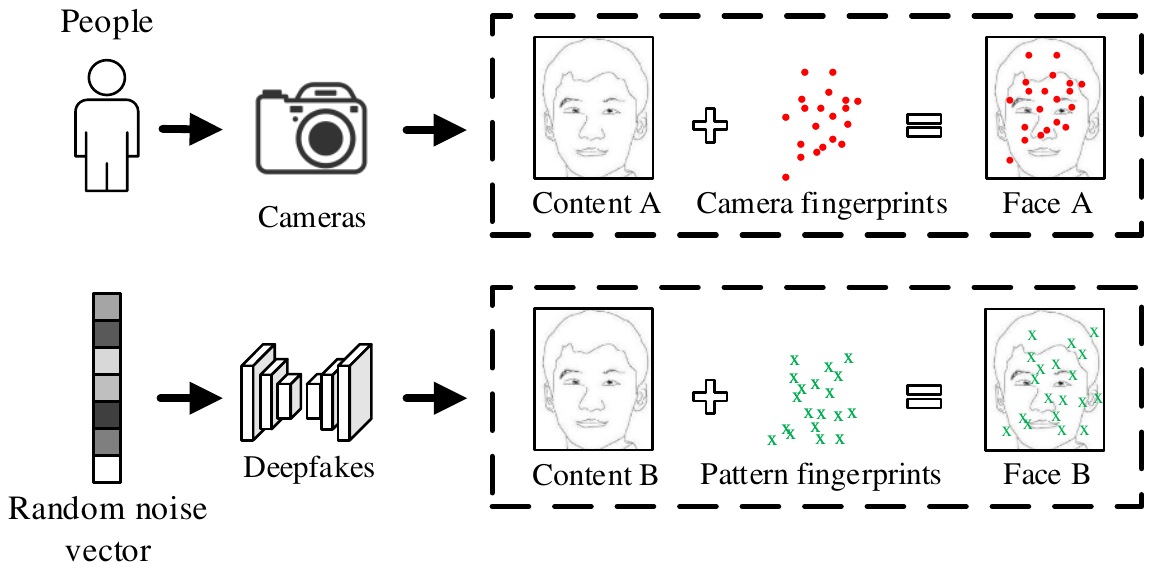}
  \caption{The formation of face images by image capturing and Deepfakes.}\label{show_sources}
\end{figure}

\begin{figure}
  \centering
  \includegraphics[width=3.4in]{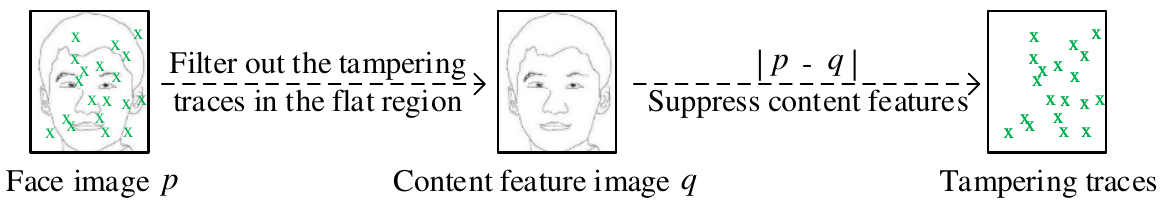}
  \caption{Illustrations of the forensic clue extraction.}\label{show_traces}
\end{figure}

Though Deepfake can create scarily real face images from scratch, there are still some intractable difficulties for the Deepfake-generated face image forgeries to keep global illumination and geometry consistency. This can be alleviated by either explicitly modelling or implicitly learning from the data, yet an imprecise estimation or an improper regularization to the training cost might lead to some subtle manipulation traces in face image regions. Fig. \ref{show_sources} compares the formation processes of face images by camera capturing and Deepfakes.
For fake face images generated by Deepfakes, it has been claimed that there exist unique pattern fingerprints or manipulation traces that are specific to different models \cite{DoGANs,Fingerprints}. That is, subtle manipulation traces are possible clues to detect Deepfake face images, and the key issue is to preserve the traces while suppressing image content. Fig. \ref{show_traces} illustrates the ideal process of extracting subtle manipulation traces for forensics. If image content can be preserved well, we can obtain the traces by subtracting image content from the original image. We observe that the guided filter, which is an explicit image filter derived from a local linear model \cite{guided_filter}, can serve as an edge-preserving image smoothing operator. Like the popular bilateral filter, the guided filter has better behavior near edges and faster speed. Motivated by its successes in many applications such as noise reduction and haze removal, we exploit the guided filter to preserve image content for better residual extraction. Specifically, instead of predicting residuals from local pixel relationships, the residuals are directly extracted by subtracting image content from the original image.
The residuals are referred to be the ``guided residuals". This is a fine-grained residual extraction method for highlighting manipulation traces.

As claimed earlier, most existing approaches exploit spatial or residual features to detect the Deepfake. The residual features often improve detection accuracies for high-quality images, yet help little for low-quality images. The reason behind this is that there are usually some post-processing operations such as lossy compression and resizing for low-quality images, which launder the manipulation traces \cite{FaceForensics++}. However, this is usually inevitable for almost all images and videos spreading over social media. For Deepfake detection, we must consider complex Internet scenarios where face images might have low qualities. Actually, spatial and residual features are correlated with each other. The spatial features that extracted from RGB images usually provide more discriminative information for low-quality images, and the residual features, which characterize well the manipulation traces left by Deepfake forgeries, are suitable for high-quality images. For Deepfake detection under complex Internet environment, adaptive feature fusion is an effective way to improve detection accuracy and robustness.

In this work, we propose a guided residuals network (GRnet), to detect the Deepfake face forgeries. It distills the above insights into a dual-stream model. Specifically, RGB images and guided residual images are fed into the backbone network to learn spatial and residual features, respectively. Moreover, an attention fusion mechanism (AFM) is presented for feature fusion, which effectively fuses the learned features in a mutually reinforcing way by adaptively allocating the weights in terms of the cross entropy loss values of two streams. The main contributions of this work are three-fold.
\begin{itemize}
  \item Instead of predicting residuals, we propose a novel manipulation trace extractor (MTE) by introducing the guided filter, which preserves well subtle manipulation traces while suppressing image content. It expands the applications of the guided filter, and overcomes the potential bias in the prediction-based residuals.
  \item An effective AFM is designed for feature fusion, which adaptively allocates the weights to spatial and residual features in terms of the cross entropy loss values of two streams. It also uses the channel attention module to establish the dependency relationship between manipulation traces.
  \item We propose a dual-stream GRnet, which learns both spatial-domain features and residual-domain features, for the Deepfake detection under complex Internet scenarios. It robustly detects fake face images with either high-qualities or low-qualities. The experimental results on four public fake face datasets prove that the proposed GRnet achieves better accuracy and robustness than the state-of-the-art works.
\end{itemize}

\indent The remainder of this paper is organized as follows. Section \uppercase\expandafter{\romannumeral2} briefly introduces the related work. Section \uppercase\expandafter{\romannumeral3} presents the GRnet model. Section \uppercase\expandafter{\romannumeral4} reports the experimental results with some analysis. Conclusion is made in Section \uppercase\expandafter{\romannumeral5}.

\section{Related Work}
\subsection{Face Image Forgeries}
In recent years, many works were presented for face image forgeries \cite{faceswap_tmm}. Among them, some works such as Glow \cite{glow} and GANimation \cite{GANimation} were proposed for facial expression manipulation. Glow is a generative flow using the invertible $1\times1$ convolution, which proves that the generative model optimized towards the plain log-likelihood objective can efficiently synthesize face images with realistic-looking facial expressions. GANimation is a GAN conditioning scheme based on action unit annotations, and its attention mechanism makes it be robust to background and lighting conditions. The other works including BEGAN \cite{BEGAN}, PGGAN \cite{pggan} and StyleGAN \cite{stylegan} can synthesize hyper-realistic fake face images directly, or change facial attributes and styles such as gender, age, and hair color \cite{StarGAN}.
PGGAN, which grows the generator and discriminator progressively, can synthesize fake face images with the spatial resolution up to 1024$\times$1024. By borrowing the idea of style transfer, StyleGAN proposes an alternative generator architecture for GAN, which leads to an automatically learned, unsupervised separation of high-level attributes and stochastic variation in generated images. For facial video forgeries, there also exist some approaches such as FaceSwap \footnote{https://github.com/MarekKowalski/FaceSwap.}, DeepFakes \footnote{https://github.com/deepfakes/faceswap.}, Face2Face \cite{Face2Face} and NeuralTextures \cite{NeuralTextures}, which replace faces, and even animate facial expressions of the target video from the source actor.

These above-mentioned works achieve photo-realistic face image forgeries, but they might still leave some manipulation traces. Unlike modeling biological signal inconsistencies (such as eye blinking\cite{eye_blinking} and head poses\cite{head_poses}), implicitly learning features from inherent manipulation traces is a more general way, which can detect various face image forgeries.

\subsection{Face Forensics Works}
For the early AI-based face image forgeries, there are still some noticeable artifacts between a real face and a fake one, which include asymmetrical faces, bad teeth and weird ears. Matern et al. \cite{visual_artifacts} exploited the visual artifacts such as imprecise geometry and reflection details to expose fake face images. To expose fake videos generated by Deepfake, Li et al. \cite{eye_blinking} exploited the inconsistency of eye blinking to design hand-crafted features. Later, deep learning was introduced into the detection of the face image forgeries \cite{SimulateArtifacts}. Dang et al. \cite{ImbalanceData} designed a customized convolutional neural network (CNN) to extract features from a tampered region, which can detect face images manipulated by BEGAN and PGGAN. Due to the lack of global constraints, Deepfake-generated face images and real faces have different facial part configuration. Yang et al. \cite{LandmarkSVM} presented to expose Deepfake face images by using landmark locations. Li et al.\cite{binary_lihaodong} exploited the disparities in color components to expose fake face images. To detect facial video forgeries, Afchar et al.\cite{MesoNet} presented a compact MesoNet, in which a low number of layers are focusing on the mesoscopic properties of images.

In addition, there are several works that exploit residual-domain features. Zhou et al. \cite{SRMfilter} proposed a two-stream Faster R-CNN for tampering region detection, in which the Spatial Rich Model (SRM) for image steganalysis is exploited to learn residual-domain features. Inspired by the concept of residuals, Bayar et al. \cite{constrainedCNN} developed a constrained convolutional layer, which jointly suppress image content and highlight manipulation traces, to learn residual-domain features for general image forensics. To expose fake face images generated by PGGAN, Mo et al. \cite{binary_highpassfilter} proposed to transform the original images into the residual images via high pass filter, which are fed into a carefully-designed CNN for binary classification. We also proposed an Adaptive Manipulation Traces Extraction Network (AMTEN), which exploits an adaptive convolution layer to predict residuals \cite{AMTEN}. The residuals are reused in subsequent layers to maximize manipulation traces by updating weights during back-propagation pass.

In essence, the existing works extract manipulation traces from residuals via a predictor, which may bring potential bias due to the prediction method. In this paper, we introduce the guided filter to protect image content. Then, the manipulation traces are highlighted by subtracting the content features from the original image, which avoids the prediction bias. However, learning features from residual-domain also has its own limitations and drawbacks. That is, the manipulation traces in low-quality face images might have been destroyed, exploiting only residual-domain features easily leads to poor detection accuracy. To address this issue, we fuse the features learned from spatial stream and residual stream to improve detection accuracy and robustness when dealing with high-quality and low-quality face images.

\subsection{Feature Fusion}
For image forensics, fusing two or more types of features feature can improve detection accuracy and robustness \cite{fusion_tmm}. The existing feature fusion strategies mainly include sum, max, min and concatenation \cite{sum_fusion,fusion_strategy}. In recent years, Bahdanau et al. \cite{att_first} proposed an attention-based fusion for machine translation. Later, the attention based feature fusion has attracted wide attentions in various computer vision tasks such as image classification \cite{att_classification}, scene segmentation \cite{att_segmentation} and fine-grained image recognition \cite{att_recognition}.
Wang et al. \cite{non_local_att} proposed a non-local attention mechanism to capture long-range dependencies for video classification tasks. Huang et al. \cite{CrissCross_att} proposed a criss-cross attention module to capture contextual information for semantic segmentation. Fu et al. \cite{att_segmentation} proposed a dual attention network to adaptively integrate local features with their global dependencies based on the self-attention mechanism. Zhu et al. \cite{ARnet} also presented an adaptive attention and residual refinement network, namely AR-Net, for copy-move forgery detection.

However, the above attention mechanisms are only used in single-stream networks for better feature representation, without considering the fusion between multi-stream features. For face forensics, both spatial features and residual features have their own advantages and disadvantages. In fact, they are complementary to each other.
In this work, an AFM is proposed to enable the CNN model selectively learn more residual or spatial features when dealing with high-quality and low-quality face images, respectively. In addition, we exploit the channel attention module to
enhance feature representation capability. Specifically, AFM treats different types of features unequally. Thus, it provides additional flexibility for feature fusion when dealing with different types of features, which enables final features be more discriminative for face forensics tasks under complex scenarios.

\section{Methodology}

\begin{figure*}
  \centering
  \includegraphics[width=6.5in]{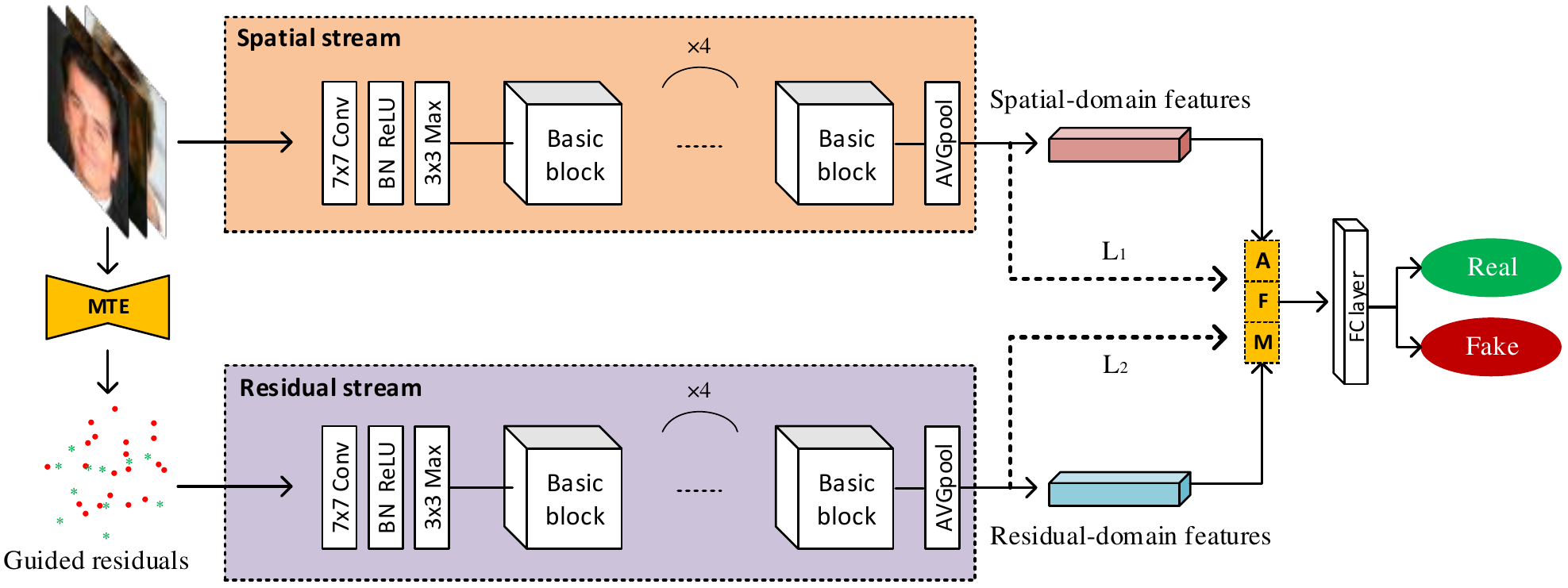}
  \caption{Overview of the proposed framework for detecting fake faces.}\label{architecture}
\end{figure*}

\subsection{Overview}
\subsubsection{Complementarity Analysis}
For image forensics tasks, both spatial-domain features and residual-domain features are either hand-crafted or implicitly learned from image data in an end-to-end manner. In general, spatial-domain features contain rich manipulation traces for image forensics, but they might also contain image content information that are suitable for image classification and recognition. However, image content information does not benefit to and even has negative impact on image forensics. As we know, image forensics is to detect image forgeries by exposing their manipulation traces that are independent of image content. Luckily, implicitly learning features from residuals can suppress the side effects of image content. Nevertheless, the manipulation traces left by image forgeries might have been partially lost in the residuals when suppressing image content.
Especially when candidate images have low visual qualities, the manipulation traces are partially destroyed, which easily leads to performance loss.

As analyzed above, we exploit the complementarity of spatial and residual features for image forensics. Specifically, for low-quality input images, more spatial-domain features, which contain rich texture information, are exploited to remedy the deficiencies of the residual-domain features. For high-quality images, more residual-domain features, which can alleviate the side effects of irrelevant image content, are exploited, simply because they are more discriminative than the spatial-domain features, especially when they are learned from manipulation traces via CNNs. Moreover, the guided residuals capture the manipulation traces much better than the previous prediction based residuals. Thus, the guided residuals are exploited for learning residual-domain features.

\subsubsection{Dual-stream Architecture}
In this work, our main goal is to identify fake face images under complex Internet scenarios. To enhance the detectors' robustness when dealing with face images with different visual qualities, a dual-stream GRnet is proposed, which learns spatial-domain features and residual-domain features in a mutually reinforcing way. Fig. \ref{architecture} is the framework of the GRnet.

Given an input RGB face image, we first convert it into a guided residual image by MTE. Then, both RGB images and guided residual images are fed into the backbone network for feature learning, respectively. Since the CNN based approaches are data hungry and small training data might lead to over-fitting, pre-training the CNN model is an effective way to address this issue. In this work, the ResNet-18 \cite{ResNet}, which has been pre-trained on the ImageNet dataset\cite{imagenet}, is exploited as the backbone network for two streams. Since the ResNet-18 is only used for feature learning in our framework, we remove full connection layers from it. The output of the bottleneck layer for each stream is a 512-dimension feature.
To improve the detectors' robustness under different scenarios, the spatial and residual features learned from two streams are fused via the AFM to emphasize the relationship between feature channel maps and adaptively adjust the weights for them.

To formulate the proposed GRnet, a quadruple variable $M$ is introduced. That is,
\begin{equation}\label{eq_1}
     M = (S_{rgb}, S_{gr}, F, D)
\end{equation}
where $S_{rgb}$ and $S_{gr}$ represent the spatial stream and the residual stream, respectively. $F$ is the AFM and $D$ denotes the detector. The learned features $f_{rgb}$ and $f_{gr}$ from each stream should have the same dimension, which are input into the AFM for feature fusion. The fused feature $F_{attention}$ is expressed as
\begin{equation}\label{eq_2}
     F_{attention} = F(f_{rgb}, f_{gr})
\end{equation}
Thus, GRnet can be formulated as an optimization problem. That is,
\begin{equation}\label{eq_3}
     \mathop{\min} \frac{1}{N} \sum_{i=1}^{N} L[D(F(f_{rgb}, f_{gr})), y]
\end{equation}
where $N$ is the number of samples, $L$ and $y$ represent the loss function and the one-hot encoding label vector, respectively.

\subsection{Manipulation Trace Extractor}
Up to now, existing works use either a fixed predictor \cite{pip1popescu} or a learning-based predictor \cite{constrainedCNN} to obtain prediction residuals. That is, the residuals are obtained by subtracting the predicted pixel values from the original pixel values. Let $p$ be the input image and $P(\cdot)$ be the predictor. The prediction residuals $R$ can be expressed as
\begin{equation}\label{eq_4}
     R = P(p) - p
\end{equation}

For image forensics, many works learn statistical features from the prediction residuals \cite{Forensics6fridrich2012,SPAM,MedResKang}. Thus, accurate residuals are critical for image forensics. To avoid the bias issue for the prediction based residuals, we suppress image content and highlight manipulation traces by introducing the guided filter.

The guided filter is an edge-preserving smoothing operator, which computes the output by the content of a guidance image $I$ \cite{guided_filter}. The guidance image should be the input image itself or another arbitrary image.
In our method, the guidance image uses the input face image to preserve the content features and filter out the manipulation traces in the flat region. The key assumption behind the guided filter is a local linear model between the guidance image $I$ and the output image $q$.
We assume that $q$ is a linear transform of $I$ in a window $\omega_k$ centered at the pixel $k$. That is,
\begin{equation}\label{eq_5}
     q_i = a_k I_i + b_k, \forall i \in \omega_k
\end{equation}
where $a_k$ and $b_k$ are some linear coefficients assumed to be constants in $\omega_k$. In reference \cite{guided_filter}, their values are defined as
\begin{equation}\label{eq_6}
     a_k = \frac{\frac{1}{|\omega|} \sum_{i \in \omega_k}I_i p_i - \mu_k \bar{p}_k}{\sigma_{k}^{2}+\epsilon}
\end{equation}
\begin{equation}\label{eq_7}
     b_k = \bar{p}_k - a_k \mu_k
\end{equation}
where $|\omega|$ is the number of pixels in $\omega_k$, $\bar{p}_k$ is the mean of $p$ in $\omega_k$, $\sigma_{k}^{2}$ and $\mu_k$ are the variance and mean of $I$ in $\omega_k$, respectively. $\epsilon$ is a regularization parameter. After obtaining the linear coefficients $\{a_k, b_k\}$, we can compute the filtering output image $q_i$ by Equation (5).

However, for a pixel $i$ that is involved in all the overlapping windows $\omega_k$, its output $q_i$ in Equation (5) is not identical when it is computed in different windows. A simple strategy is to average all the possible values of $q_i$. That is, after computing $\{a_k, b_k\}$ for all the windows $\omega_k$ in the image, the filtering output $q_i$ is obtained as follows.
\begin{equation}\label{eq_8}
     q_i = \frac{1}{|\omega|} \sum_{k|i \in \omega_k}(a_k I_i + b_k)
\end{equation}
Let $\overline{a}_i$ and $\overline{b}_i$ be the average coefficients of all windows overlapping $i$. Equation (8) can be rewritten as
\begin{equation}\label{eq_9}
     q_i = \bar{a}_i I_i + \bar{b}_i
\end{equation}

\begin{figure}
  \centering
  \includegraphics[width=3.4in]{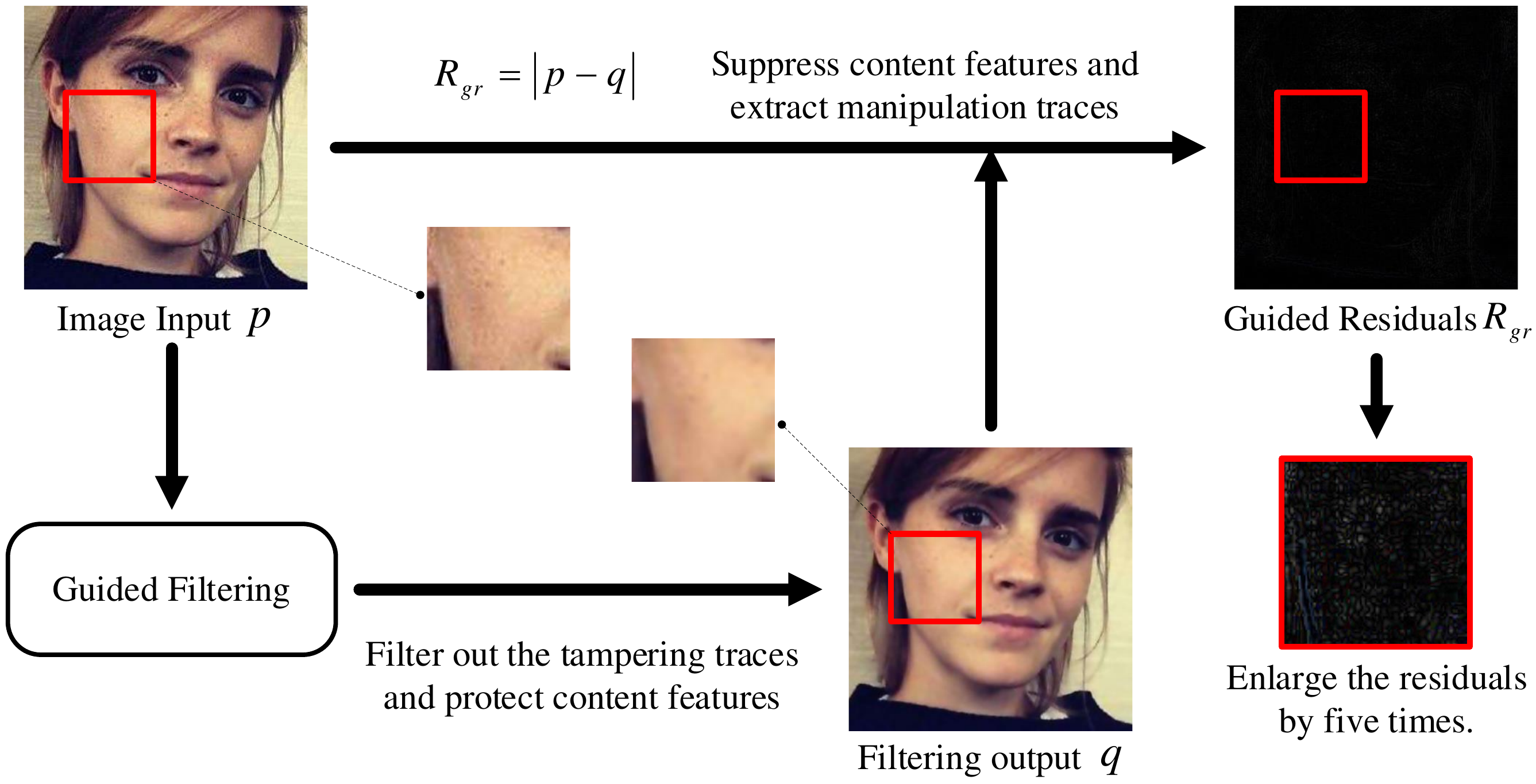}
  \caption{Manipulation trace extractor. To facilitate observation, the guided residuals $R_{gr}$ in the red box are enlarged five times.}\label{guided_res_process}
\end{figure}

As claimed earlier, a forged face image can be viewed as the combination of content features $C$ and manipulation traces $T$ (see Fig. \ref{show_sources}). Thus, the input face image $p$ can be simply expressed as
\begin{equation}\label{eq_10}
     p = C + T
\end{equation}
The local linear model behind the guided filter ensures that the filtering output $q$ preserves well image content features $C$ and removes the manipulation traces $T$ in the flat region. Thus, the filtering output $q$ of the input image $p$ is equal to $q = C$. Then, the guided residuals $R_{gr}$ is obtained by subtracting the filtering output $q$ from the input image $p$. Fig. \ref{guided_res_process} shows the extraction of the guided residuals. Note that we take the absolute value to avoid possible negative value. That is,
\begin{equation}\label{eq_11}
     R_{gr} = |p - q| = T
\end{equation}

\begin{figure*}
  \centering
  \includegraphics[width=6.5in]{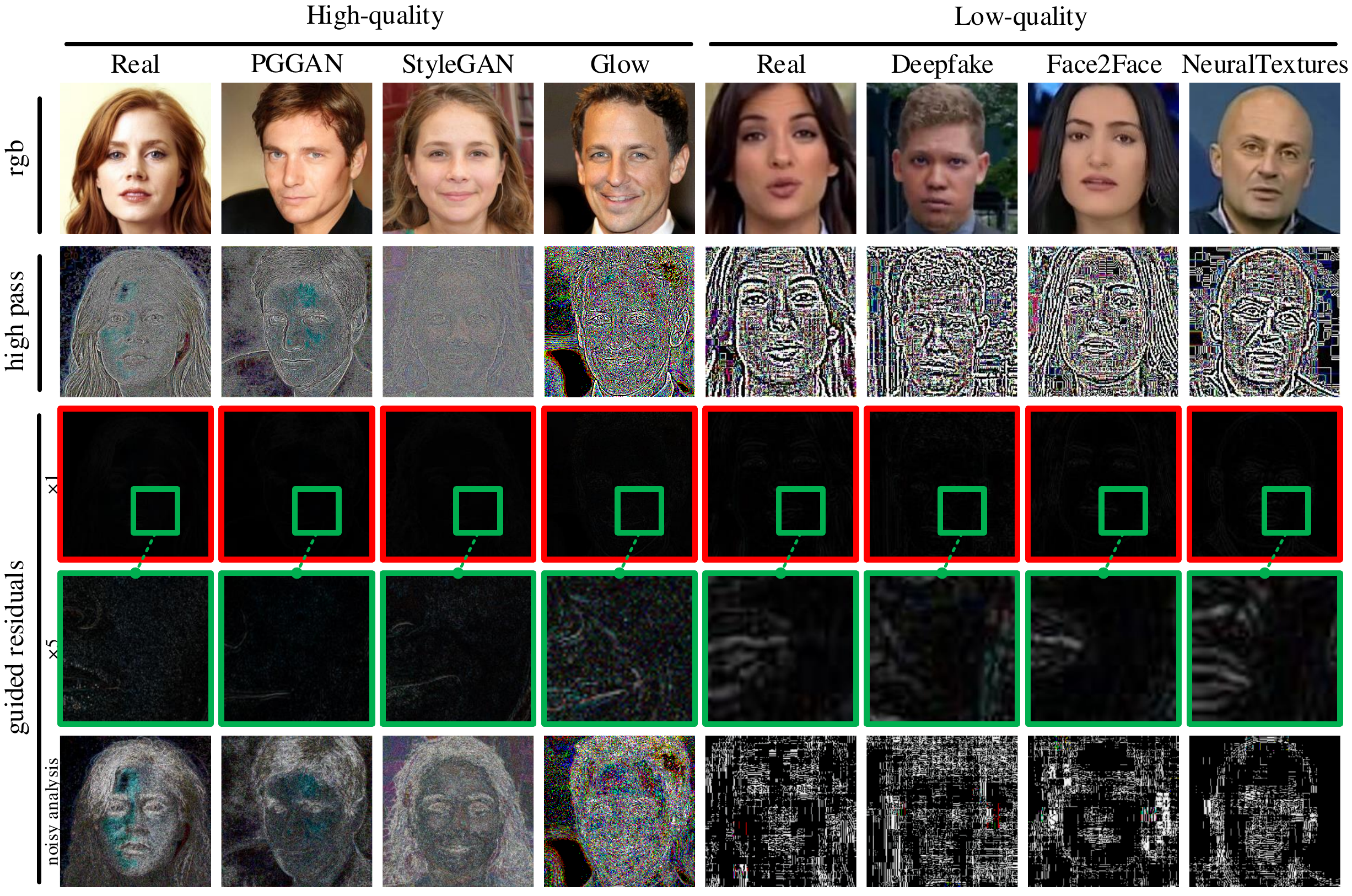}
  \caption{Comparison of prediction residuals and guided residuals. To highlight the differences of the guided residuals among  different facial forgery techniques, the texture details in the green boxes are enlarged  by 5 times, and noisy analysis is made for the guided residual images in the red boxes.}\label{res_compare}
\end{figure*}

Equation (4) is a general pipeline for extracting prediction-based residuals. The guided filter can be treated as a fixed predictor, but it does not exploit the relationship between local pixels for prediction.
That is, the guided filter is used to preserve content features and remove manipulation traces. Then, we suppress the content features and extract the manipulation traces (i.e. guided residuals). Thus, this is an intuitive yet more suitable way to obtain residuals.
Fig. \ref{res_compare} compares the residuals obtained by high-pass filter \cite{binary_highpassfilter} and MTE for both high-quality and low-quality face images. The first row shows the color images, including real face images and tampered images. The second row shows the prediction residuals obtained by high-pass filter, in which there are still some details about image content. The third and fourth rows show the guided residuals, in which most image contents are removed. In addition, we further enlarge the guided residuals by noise analysis \footnote{https://29a.ch/photo-forensics/\#noise-analysis}, and the resultant guided residuals are shown in the fifth row.
From Fig. \ref{res_compare}, we can observe that for high-quality input images, the residuals have rich micro-texture details, and there are obvious differences among the residuals of different forged images. This is very beneficial for the forensics model to learn discriminative features. For the low-quality input image, we can see from the noise analysis in the fifth row that there are many white blocky textures in the residual images.
These similar textures make it difficult for the forensics model to discriminate the differences among different forged images. This also proves that the manipulation traces in low-quality images are laundered.

\subsection{Attention Fusion Mechanism}
\begin{figure}
  \centering
  \includegraphics[width=3.4in]{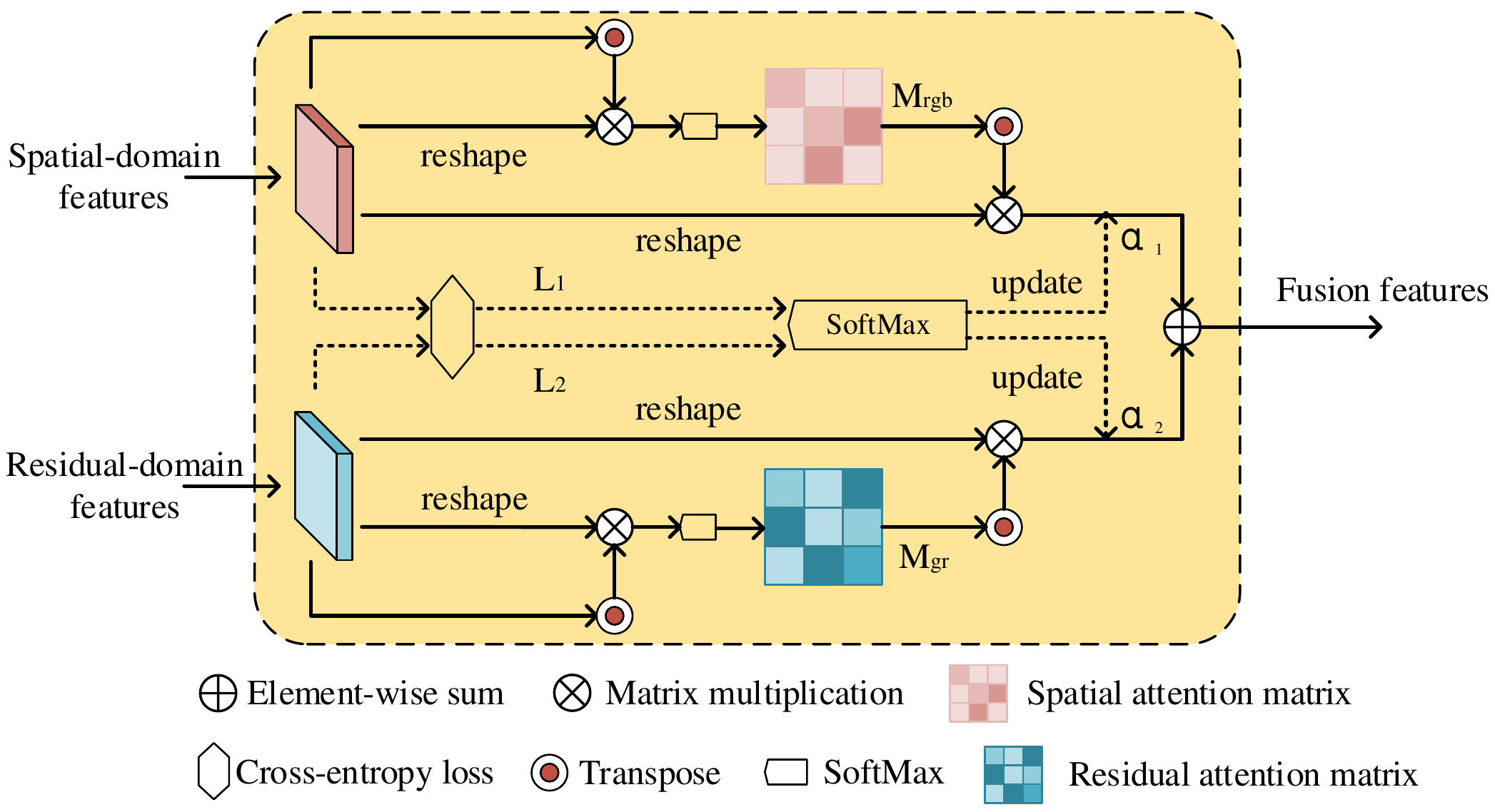}
  \caption{Attention fusion mechanism. $L_{1}$ and $L_{2}$ are the cross-entropy loss of two streams, respectively.}\label{AFM}
\end{figure}

Feature fusion refers to combining two or more feature vectors to form a single feature vector, which should be more discriminative than any of the input feature vectors. For image forensics, feature fusion improves both detection accuracy and robustness. In this work, both spatial and residual features are learned by the dual-stream GRnet. For each stream, a 512D feature channel map, which can be treated as the response to the manipulation traces left by face forgeries, is learned. Since feature fusion is to improve feature representation capability, how to exploit the dependency between these responses is a key issue to expose the manipulation traces. The GRnet is dedicated to expose Deepfake face images spreading over Internet, which might have high or low qualities. As claimed earlier, the spatial stream is preferable for learning features from input images with low-qualities, whereas the residual stream is suitable for learning features from high-quality images. How to adaptively allocate the weights to two streams is the key to improve detection performance when dealing with face images with distinct visual qualities. Thus, we design an AFM, whose structure is shown in Fig.\ref{AFM}.

Firstly, the dependence between feature channel maps is exploited to improve feature representation capability. Let $f_{rgb}$ be the set of the spatial features $f_{rgb}$, where $f_{rgb} \in \mathbb{R}^{C \times H \times W}$. $C$, $H$ and $W$ are the channel, height and width, respectively. $f_{rgb}$ is reshaped into $\mathbb{R}^{C \times N_p}$, where $N_p$ is the number of pixels. Then, the reshaped features are matrix multiplied with the transpose of $f_{rgb}$, and the result is further processed with the softmax function to obtain the spatial attention matrix $M_{rgb} \in \mathbb{R}^{C \times C}$.

\begin{equation}\label{eq_11}
     m_{rgb}^{ji} = \frac{exp(f_{rgb}^{i} \cdot f_{rgb}^{j})}{\sum_{i=1}exp(f_{rgb}^{i} \cdot f_{rgb}^{j})}
\end{equation}
where $m_{rgb}^{ji}$ represents the influence of the $i^{th}$ channel on the $j^{th}$ channel. Similarly, the residuals attention matrix $M_{gr} \in \mathbb{R}^{C \times C}$ can be obtained as follows.
\begin{equation}\label{eq_12}
     m_{gr}^{ji} = \frac{exp(f_{gr}^{i} \cdot f_{gr}^{j})}{\sum_{i=1}exp(f_{gr}^{i} \cdot f_{gr}^{j})}
\end{equation}
The reshaped $\{f_{rgb}, f_{gr}\}$ is also matrix multiplied with the transpose of $\{M_{rgb}, M_{gr}\}$ to obtain a new feature representation $\{f'_{rgb}, f'_{gr}\}$.

Secondly, both the spatial stream loss $L_{1}$ and the residual stream loss $L_{2}$ in Fig. \ref{AFM} are used to allocate the weights of two streams adaptively. In general, the loss value of each stream is inversely proportional to its weight. Specifically, $L_{1}$ and $L_{2}$ are fed into the softmax function to define the weights $\alpha_i$ which are subject to $\sum_{i}\alpha_i = 1$.
\begin{equation}\label{eq_13}
     \alpha_i = 1 - \frac{exp(L_{i})}{\sum_{j}exp(L_{j})}
\end{equation}
During the back-propagation pass, the weights are iteratively updated through the loss value of each epoch. 

Thirdly, we aggregate the two streams via the element-wise sum operation as follows
\begin{equation}\label{eq_14}
     F_{attention} = \alpha_1 f'_{rgb} + \alpha_2 f'_{gr}
\end{equation}

\section{Experimental results}

\begin{figure}
  \centering
  \includegraphics[width=3.0in]{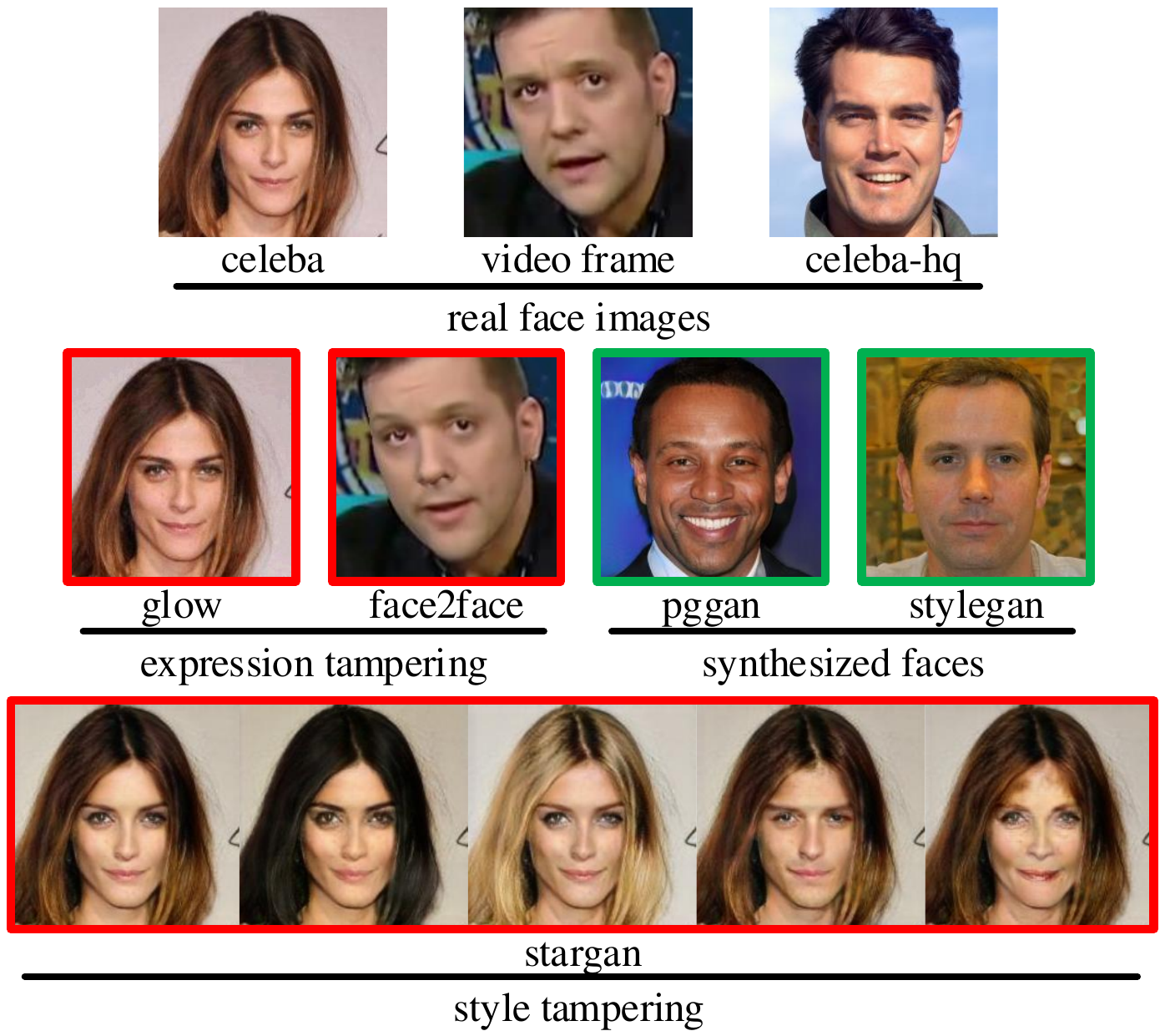}
  \caption{Some face images from the HFF dataset. In the 2nd row, the face images marked with red boxes are forged by expression tampering techniques, and the face images marked with green boxes are artificially synthesized by GANs. In the 3rd row, facial styles are changed, including hair colors, gender and age.}\label{HFFD}
\end{figure}

\subsection{Benchmark Datasets}
\indent The Hybrid Fake Face (HFF) \cite{AMTEN} dataset contains 155k high-quality face images, which include both real and fake face images. For the real face images, they come from the CelebA dataset \cite{celeba}, the CelebA-HQ dataset \cite{pggan} and YouTube video frames \cite{Faceforensics}, respectively. The fake face images are obtained by five typical facial forgery techniques including PGGAN \cite{pggan}, StyleGAN \cite{stylegan}, Face2Face\cite{Face2Face}, Glow \cite{glow}, and StarGAN \cite{StarGAN}.
Note that both PGGAN and StyleGAN can synthesize face images that do not exist up to the spatial resolution of 1024$\times$1024, Face2Face and Glow were presented for facial expression transferring, whereas StarGAN was proposed for changing face styles such as hair color and gender by multi-domain image-to-image translation. Some samples are randomly selected from the HFF dataset, which are shown in Fig. \ref{HFFD}.

\begin{figure}
  \centering
  \includegraphics[width=3.0in]{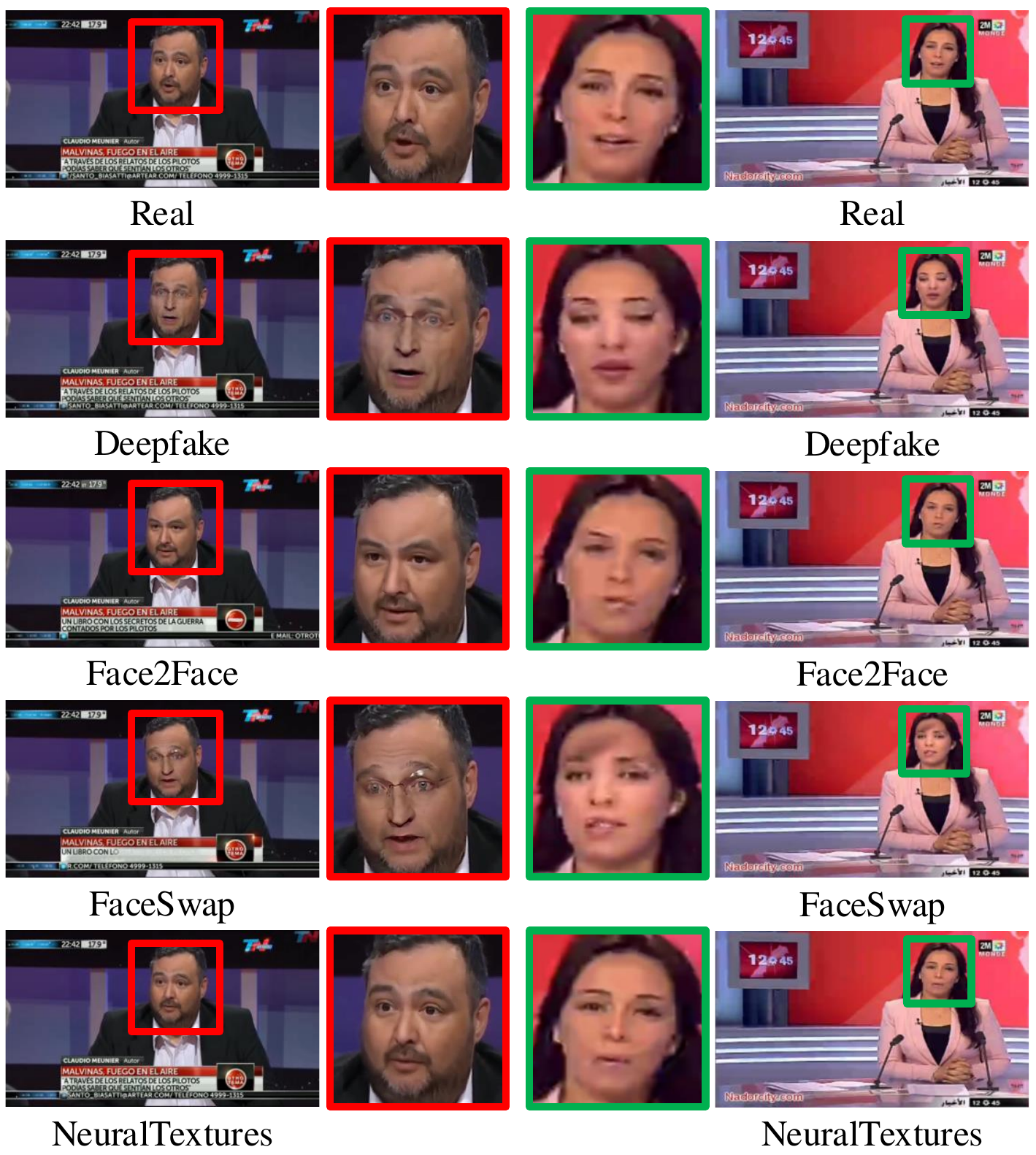}
  \caption{Face video frames from the FF dataset. The red box indicates that the face image belongs to the HQ dataset and the green box indicates that the face image belongs to the LQ dataset.}\label{FFD}
\end{figure}

The FaceForensics++ (FF) \cite{FaceForensics++} dataset is made up of 1k video sequences that have been manipulated with four facial forgeries including Deepfake, Face2Face, FaceSwap and NeuralTextures, respectively. The FF dataset provides visually lossless high-quality (HQ) videos (the quantization parameter is set with a constant of 23) and visually lossy low-quality (LQ) videos (the quantization parameter is set with a constant of 40 for H.264/AVC compression), which are widely-used in social networks. In the experiment, some frames are extracted from each video with two compression levels. Then, we obtain the face images with the size of 128$\times$128 by simply removing background, as shown in Fig. \ref{FFD}. For real face images, 50k and 10k images are used as the training set and the testing set, respectively. For fake face images, we select 12.5k and 2.5k sample images as the training set and the testing set for each forgery technique, respectively.

\begin{figure}
  \centering
  \includegraphics[width=3.0in]{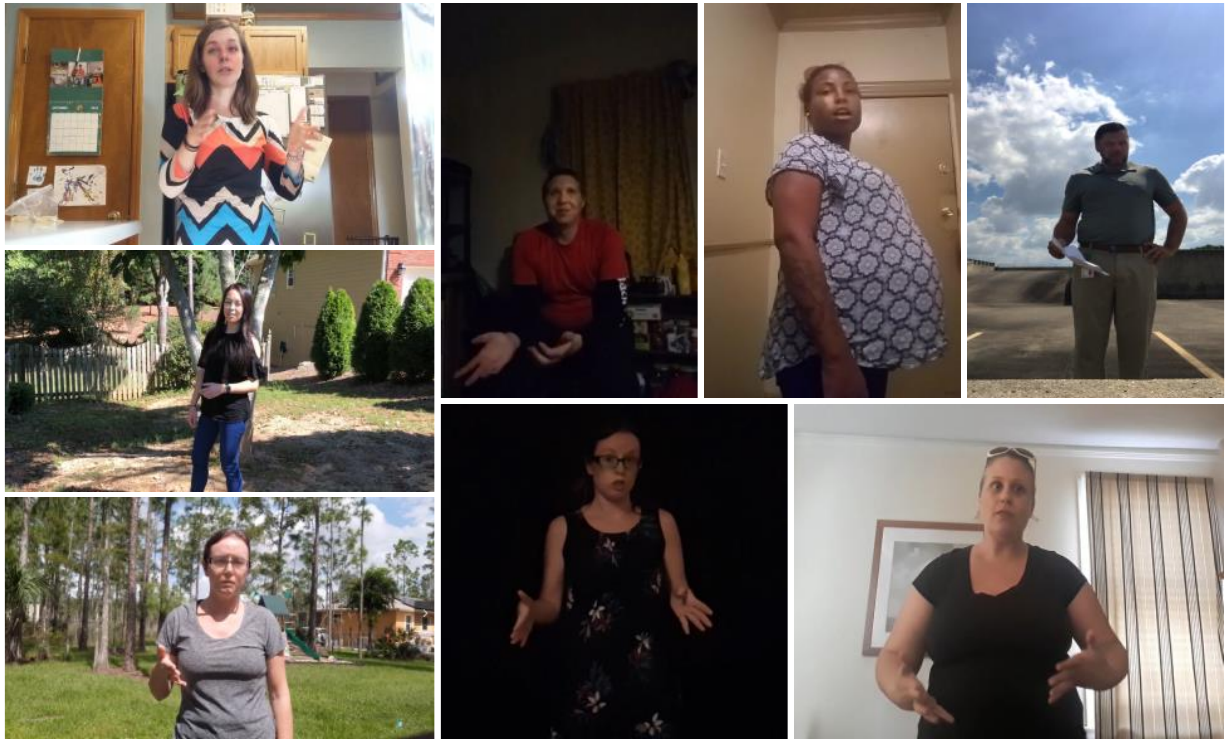}
  \caption{Face video frames from the DFDC dataset.}\label{DFDC}
\end{figure}

The DeepFake Detection Challenge (DFDC) \cite{DFDC} dataset contains more than 100k real and fake video sequences (about 470GB). There are various actors with diverse attributes such as age, gender and skin-tone. They are recorded with arbitrary background, thus bringing visual diversities. Fig. \ref{DFDC} illustrates some face images from the DFDC dataset. Note that the fake videos are manipulated by two facial forgery algorithms. Thus, a binary classification is made about the trustworthiness of the face images. Specifically, we randomly select about 6,698 video sequences. For each video sequence, face images with the size of 224$\times$224 are extracted for every thirty frames. Then, we select a maximum of 10 face images for each video. Thus, there are total 31,873 real face images and 30,945 fake face images in our experiments.

\subsection{Experimental Settings}
\subsubsection{Evaluation Criterion}
For image forensics tasks, there are two commonly-used metrics, namely accuracy rate (ACC) and area under the ROC curve (AUC), for performance evaluation. In our experiments, ACC is used to evaluate the detectors' accuracy. AUC is used to measure the performance for binary classification.

\subsubsection{Baseline Models}
Several state-of-the-art works are selected as the baseline models for experimental comparisons. They are summarized as follows.
\begin{itemize}
  \item Meso-4 \cite{MesoNet}: It mainly exploits the mesoscopic properties of face images to detect fake videos generated by Deepfake.
  \item MesoInception-4 \cite{MesoNet}: It is another deep learning based work that has much better performance than Meso-4.
  \item HighPass \cite{binary_highpassfilter}: It extracts residuals via high pass filters. We exploit the high pass filter that achieves the best detection accuracy for comparisons in our tasks.
  \item MISLnet \cite{constrainedCNN}: A constrained convolution layer, which is equivalent to an adaptive residual predictor, is designed. MISLnet is a pioneer work towards universal image forgery detection. It also serves as a universal detector here for experimental comparisons.
  \item XceptionNet \cite{Xception}: It is originally presented for image classification. However, it is exploited for facial forgery detection and achieves desirable performance in \cite{FaceForensics++}.
  \item Capsule \cite{CapsuleV2}: It extends the capsule networks to expose various kinds of spoofs such as replay attack and computer generated images/videos.
  \item AMTENnet \cite{AMTEN}: It presents an adaptive manipulation traces extraction network (AMTEN) for facial forgery detection, which achieves the state-of-the-art performance.
\end{itemize}

\subsubsection{Implementation Details}
The proposed GRnet is implemented under the PyTorch framework. We use one Nvidia GeForce GTX 1080 Ti GPU to train the model and employ the ADAM with the decay parameters ($\beta_1 = 0.9$, $\beta_2 = 0.999$) for network optimization. An exponential learning rate ($\gamma = 0.5$) is used, and the initial learning rate is 0.0005. To improve model generalization capability and robustness, some data augmentation operations, which include horizontal flip, rotation, random perspective and normalization, are adopted to produce more training data from the original data.

\subsection{Ablation Study}
In this work, both MTE and AFM are two innovations for GRnet. Thus, we conduct ablation study on GRnet.
To verify the detectors' robustness under different scenarios, the same post-processing operations are applied to the HFF dataset following the reference \cite{AMTEN}. Specifically, the quality factor of JPEG compression is set to 60 (JP60), which can launder manipulation traces by weakening image quality. The kernel size of the mean filtering is set to 5$\times$5 (ME5), which smoothes image details including the manipulation traces due to image blur.

From Table I, we can observe that when only MTE is equipped, the network can achieve the best detection accuracy on high-quality data (Raw), which also proves that residual features are very good at capturing manipulation traces in high-quality images. For low-quality data (JP60 and ME5), when MTE and AFM are equipped to GRnet, the best detection accuracy is achieved. Especially in the compression scenario, MTE and AFM are equipped in the backbone network, which significantly improves the accuracy by 2.07\%. It also proves that the fusion of residual-domain and spatial-domain features can effectively improve the robustness of the network in complex scenarios.

\begin{table}[]
\centering
\footnotesize{TABLE I}
\caption*{\centering{\scriptsize{ABLATION STUDY FOR GRNET.}}}
\renewcommand{\arraystretch}{1.1}
\resizebox{80mm}{!}{
\begin{tabular}{|c|c|c|c|c|c|}
\hline
Model                  & MTE              & AFM              & ACC(Raw)       & ACC(JP60)      & ACC(ME5)       \\ \hline
\multirow{4}{*}{GRnet} & \textbackslash{} & \textbackslash{} & 99.21          & 94.41          & 98.11          \\ \cline{2-6}
                       & $\checkmark$     & \textbackslash{} & \textbf{99.98} & 94.85          & 98.89          \\ \cline{2-6}
                       & \textbackslash{} & $\checkmark$     & 99.65          & 94.97          & 98.96          \\ \cline{2-6}
                       & $\checkmark$     & $\checkmark$     & 99.96          & \textbf{96.48} & \textbf{99.15} \\ \hline
\end{tabular}
}
\end{table}

\subsection{Key Module Comparison}
\subsubsection{Comparison for Residuals}
In existing works, there are several common residual extraction methods, which include the fixed predictors (high pass filter \cite{binary_highpassfilter} and the SRM filter \cite{SRMfilter}) and the learning-based predictors (AMTEN \cite{AMTEN} and constrained convolution layer \cite{constrainedCNN}). They are selected as the baseline methods to make some comparisons with MTE. For fair comparison, we compare the detection accuracies when using the same backbone networks according to the experimental benchmark in \cite{AMTEN}. The HFF dataset is selected for experiments.

\begin{table}[]
\centering
\footnotesize{TABLE II}
\caption*{\centering{\scriptsize{COMPARISON FOR VARIOUS TYPES OF RESIDUAL METHODS}}}
\renewcommand{\arraystretch}{1.1}
\resizebox{80mm}{!}{
\begin{tabular}{|c|c|c|}
\hline
Styles                                                                              & Residual Methods              & ACC              \\ \hline
\multirow{2}{*}{Fixed predictor}                                                    & High pass filter              & 97.50\%          \\ \cline{2-3}
                                                                                    & SRM filter                    & 97.49\%          \\ \hline
\multirow{2}{*}{\begin{tabular}[c]{@{}c@{}}Learning-based\\ predictor\end{tabular}} & Constrained convolution layer & 95.24\%          \\ \cline{2-3}
                                                                                    & AMTEN                         & 98.52\%          \\ \hline
Non-prediction                                                                      & MTE                           & \textbf{99.04\%} \\ \hline
\end{tabular}
}
\end{table}

Table II reports the detection accuracies when using different residual extraction methods. Specifically, high pass filter and the SRM filter, as the fixed predictors, achieve the accuracy rates of 97.50\% and 97.49\%, respectively.
For the learning based predictors, the constrained convolution layer achieves an accuracy of 95.24\%, and AMTEN achieves an accuracy up to 98.52\%. The constrained convolution layer works by resetting specific coefficients in the convolutional kernel after each iteration, whereas AMTEN adaptively updates the coefficients in the convolutional kernel after each iteration. Thus, AMTEN obtains better residuals than the constrained convolution layer, resulting an accuracy improvement of 3.28\%.
Unlike the fixed predictors and the learning-based predictors, the MTE achieves an accuracy rate up to 99.04\%, which is the highest. We can observe that our method achieves better accuracies than the fixed predictors and the learning-based predictors. The inherent reason behind this is that the guided residuals avoid the potential bias brought by the prediction-based methods, which characterize well the traces left by various face image forgeries.

\begin{figure}
  \centering
  \includegraphics[width=3.5in]{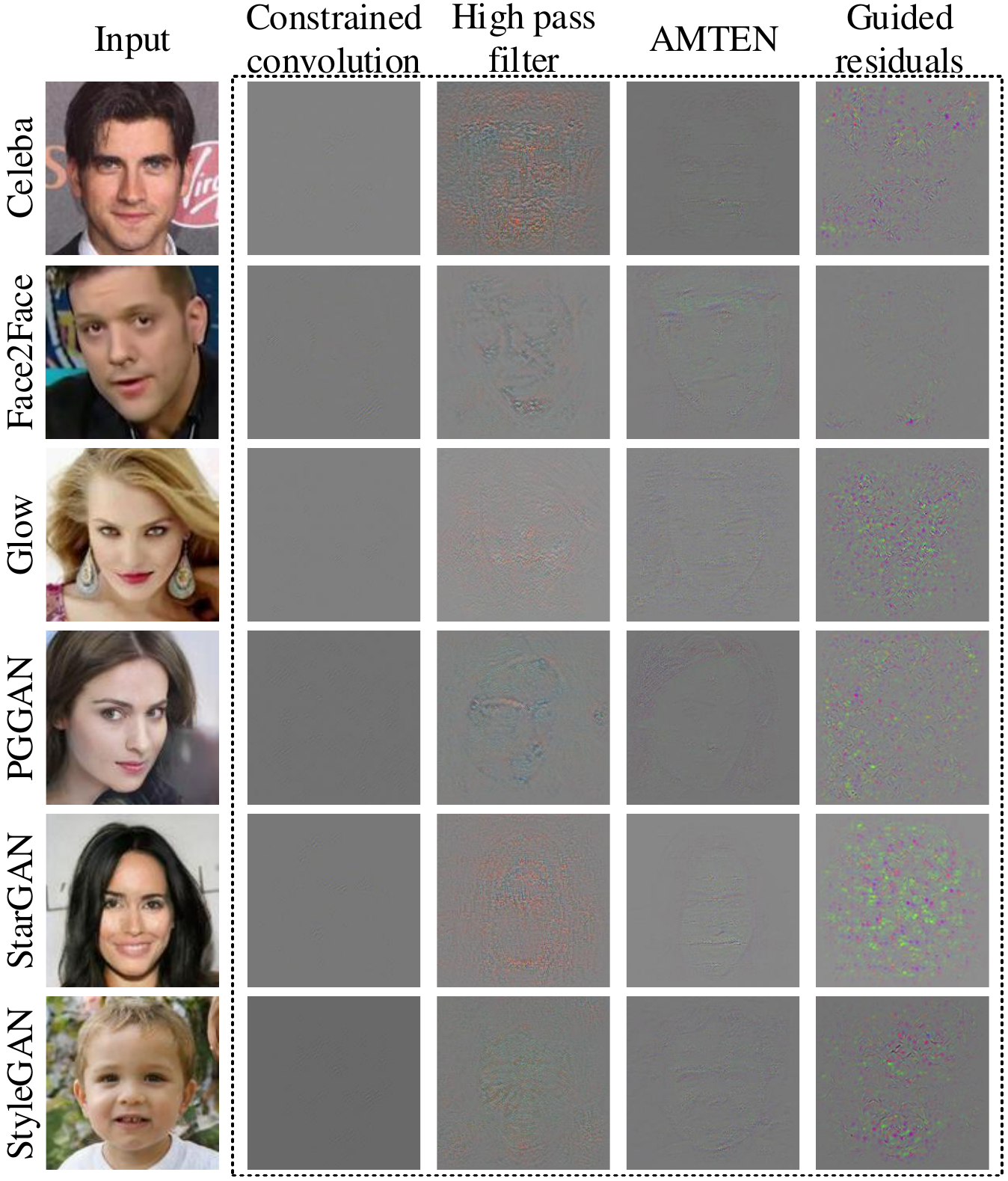}
  \caption{Visualization of features obtained by CNN detectors with different residuals.}\label{GuidedBP}
\end{figure}

To facilitate the understanding of the guided residuals obtained by MTE, we also visualize the features learned by different pre-processing methods \cite{visualization}, as shown in Fig. \ref{GuidedBP}. Though the constrained convolution layer greatly suppresses image content, the manipulation traces left in the residuals are fragile. For high pass filter and AMTEN, there are still some face contour information preserved. However, compared with the constrained convolution layer, the residuals obtained by high pass filter and AMTEN reflects better the manipulation traces. Thus, high pass filter and AMTEN achieves better detection accuracies, which are 97.50\% and 98.52\% respectively. From the fifth column of Fig. \ref{GuidedBP}, we can also observe that the guided residuals almost completely suppress image content, and thus the CNN model learns more discriminative manipulation trace features from the residuals.

\begin{table}[]
\centering
\footnotesize{TABLE III}
\caption*{\centering{\scriptsize{COMPARISON FOR VARIOUS TYPES OF FUSION METHODS}}}
\renewcommand{\arraystretch}{1.1}
\resizebox{50mm}{!}{
\begin{tabular}{|c|c|}
\hline
Fusion methods       & ACC     \\ \hline
Max fusion           & 94.61\%    \\ \hline
Concatenation fusion & 95.32\% \\ \hline
Min fusion           & 95.72\% \\ \hline
Sum fusion           & 94.85\% \\ \hline
AFM                  & \textbf{96.48\%} \\ \hline
\end{tabular}
}
\end{table}

\subsubsection{Comparison of Feature Fusion Methods}
AFM is designed for feature fusion to further improve the detection performance of the model. To prove the effectiveness of the AFM, we also conduct some experiments by comparing it and four common feature fusion strategies, namely max, min, sum and concatenation.
To better show the performance advantages of our method, the experiments are made on compressed HFF dataset. Table III reports the detection accuracies when using different feature fusion methods. From it, we can observe that AFM achieves the best accuracies of 96.48\%.

\subsection{Comparison with State-of-the-Art Approaches}
The proposed GRnet is compared with the state-of-the-art works, and the HFF, FF and DFDC datasets are selected for experiments. Among them, both HFF and FF datasets contain fake face images generated by various face manipulation techniques. For both datasets, we provide both multiple classification and binary classification. Note that multiple classification is to identify which technique is used to manipulate face images, which is a fine-grained recognition problem. For binary classification, it is used to judge the authenticity of face images, yet it has its own advantage for generalization tests.

\subsubsection{Results on the HFF Dataset}

\begin{table}[]
\centering
\footnotesize{TABLE IV}
\caption*{\centering{\scriptsize{MULTIPLE CLASSIFICATION ACCURACY RATE OF DIFFERENT FORENSICS MODELS ON HFF DATASET}}}
\renewcommand{\arraystretch}{1.1}
\resizebox{85mm}{!}{
\begin{tabular}{|l|l|l|l|l|}
\hline
\multicolumn{1}{|c|}{Methods}  & \multicolumn{1}{|c|}{Raw}   & \multicolumn{1}{|c|}{JP60}             & \multicolumn{1}{|c|}{ME5}              & \multicolumn{1}{|c|}{Average}          \\ \hline
Meso-4\cite{MesoNet}                 & 80.76\%          & 67.76\%          & 62.40\%          & 70.31\%          \\ \hline
MesoInception-4\cite{MesoNet}        & 86.40\%          & 58.68\%          & 77.68\%          & 74.25\%          \\ \hline
HighPass\cite{binary_highpassfilter} & 90.54\%          & 73.81\%          & 74.99\%          & 79.78\%          \\ \hline
MISLnet\cite{constrainedCNN}         & 93.76\%          & 86.32\%          & 79.06\%          & 86.38\%          \\ \hline
XceptionNet\cite{Xception}           & 97.17\%          & 78.62\%          & 90.88\%          & 88.89\%          \\
\hline
Capsule\cite{CapsuleV2}                & 96.75\%          & 90.02\%          & 94.44\%          & 93.74\%          \\
\hline
AMTENnet\cite{AMTEN}                   & 98.52\%          & 91.02\%          & 92.42\%          & 93.99\%          \\ \hline
\hline
GRnet                        & \textbf{99.96\%} & \textbf{96.48\%} & \textbf{99.15\%} & \textbf{98.53\%} \\ \hline
\end{tabular}
}
\end{table}

\begin{table}[]
\centering
\footnotesize{TABLE V}
\caption*{\centering{\scriptsize{BINARY CLASSIFICATION ACCURACY RATE OF DIFFERENT FORENSICS MODELS ON HFF DATASET}}}
\renewcommand{\arraystretch}{1.1}
\resizebox{85mm}{!}{
\begin{tabular}{|l|l|l|l|l|}
\hline
\multicolumn{1}{|c|}{Methods}  & \multicolumn{1}{|c|}{Raw}   & \multicolumn{1}{|c|}{JP60}             & \multicolumn{1}{|c|}{ME5}              & \multicolumn{1}{|c|}{Average}          \\ \hline
Meso-4\cite{MesoNet}                 & 76.83\%          & 62.40\%          & 63.52\%          & 67.58\%          \\ \hline
MesoInception-4\cite{MesoNet}        & 94.33\%          & 73.63\%          & 84.43\%          & 84.13\%          \\ \hline
HighPass\cite{binary_highpassfilter} & 89.06\%          & 72.97\%          & 76.88\%          & 79.64\%          \\ \hline
MISLnet\cite{constrainedCNN}         & 93.71\%          & 87.87\%          & 84.76\%          & 88.78\%          \\ \hline
XceptionNet\cite{Xception}           & 92.82\%          & 74.12\%          & 76.62\%          & 81.19\%          \\ \hline
Capsule\cite{CapsuleV2}                & 95.66\%          & 89.12\%          & 92.63\%          & 92.47\%          \\ \hline
AMTENnet\cite{AMTEN}                   & 97.66\%          & 88.91\%          & 81.36\%          & 89.31\%          \\ \hline
\hline
GRnet                        & \textbf{99.81\%} & \textbf{94.02\%} & \textbf{99.32\%} & \textbf{97.72\%} \\ \hline
\end{tabular}
}
\end{table}

Table IV reports the multiple classification accuracies when using different networks. From it, we can observe that the shallow networks including Meso-4 and MesoInception-4 can not achieve desirable results under three scenarios (Raw, JP60, and ME5), and their average accuracies are 70.31\% and 74.25\%, respectively. Among these baseline models, both XceptionNet and Capsule are deep CNNs with complex structures, and they achieve the average accuracies of 88.89\% and 93.74\% under three scenarios, respectively. HighPass exploits shallow CNN to learn features from the residuals obtained by the fixed predictor. It achieves an average accuracy of 79.78\%, which is about 5.53\% higher than that of MesoInception-4. MISLnet and AMTENnet are also shallow convolutional networks, yet their inputs are the learning based residuals. The residuals are updated iteratively during the back-propagation pass, from which more discriminative features can be learned. As a result, MISLnet and AMTENnet achieve the average accuracies of 86.38\% and 93.99\%, respectively. Note that the accuracies are close to slighter or better than that of XceptionNet. From this, we can conclude that for the deep learning based face forensics tasks, extracting the residuals as appropriate as possible can improve the feature representation capability of the CNN model.
Furthermore, the proposed GRnet combines residual-domain and spatial-domain features in a mutually reinforced way, achieving an average accuracy of 98.53\%, which is 4.54\% higher than that of the state-of-the-art work with the best performance.

Table V reports the results for the binary classification task, which are similar to the results in Table IV. From Table V, the proposed GRnet achieves an average accuracy of 97.72\%, which outperforms that of the Capsule network by a large margin up to 5.25\%. In the HFF dataset, most face images are generated by GANs that leave unique manipulation traces in the global images, instead of just manipulated local regions, which can be easily extracted by MTE. As shown in Table IV and V, the proposed GRnet has achieved much better accuracies than the baseline methods.

\begin{table}[]
\centering
\footnotesize{TABLE VI}
\caption*{\centering{\scriptsize{MULTIPLE CLASSIFICATION ACCURACY RATE OF DIFFERENT FORENSICS MODELS ON THE FF DATASET}}}
\renewcommand{\arraystretch}{1.1}
\resizebox{70mm}{!}{
\begin{tabular}{|l|c|c|c|}
\hline
\multicolumn{1}{|c|}{Methods}               & HQ               & LQ               & Average          \\ \hline
Meso-4\cite{MesoNet}                        & 52.92\%          & 50.63\%          & 51.78\%          \\ \hline
MesoInception-4\cite{MesoNet}               & 67.38\%          & 45.97\%          & 56.68\%          \\ \hline
HighPass\cite{binary_highpassfilter}        & 82.94\%          & 64.58\%          & 73.76\%          \\ \hline
MISLnet\cite{constrainedCNN}                & 82.44\%          & 65.61\%          & 74.03\%          \\ \hline
XceptionNet\cite{Xception}                  & 76.30\%          & 71.63\%          & 73.97\%          \\ \hline
Capsule\cite{CapsuleV2}                     & 97.56\%          & \textbf{97.28\%} & \textbf{97.42\%} \\ \hline
AMTENnet\cite{AMTEN}                        & 90.11\%          & 72.14\%          & 81.13\%          \\ \hline
\hline

GRnet                               & \textbf{97.79\%} & 96.58\%           &97.19\% \\ \hline
\end{tabular}
}
\end{table}

\begin{table}[]
\centering
\footnotesize{TABLE VII}
\caption*{\centering{\scriptsize{BINARY CLASSIFICATION ACCURACY RATE OF DIFFERENT FORENSICS MODELS ON THE FF DATASET}}}
\renewcommand{\arraystretch}{1.1}
\resizebox{70mm}{!}{
\begin{tabular}{|l|c|c|c|}
\hline
\multicolumn{1}{|c|}{Methods}               & HQ               & LQ               & Average          \\ \hline
Meso-4\cite{MesoNet}                        & 61.59\%          & 61.27\%          & 61.43\%          \\ \hline
MesoInception-4\cite{MesoNet}               & 64.77\%          & 77.21\%          & 70.99\%          \\ \hline
HighPass\cite{binary_highpassfilter}        & 79.23\%          & 78.79\%          & 79.01\%          \\ \hline
MISLnet\cite{constrainedCNN}                & 83.84\%          & 82.85\%          & 83.35\%          \\ \hline
XceptionNet\cite{Xception}                  & 75.17\%          & 75.98\%          & 75.58\%          \\ \hline
Capsule\cite{CapsuleV2}                       & 96.74\%          & 96.65\%          & 96.70\%          \\ \hline
AMTENnet\cite{AMTEN}                          & 85.14\%          & 84.16\%          & 84.65\%          \\ \hline
\hline

GRnet                                & \textbf{97.47\%} & \textbf{96.68\%} & \textbf{97.08\%} \\ \hline
\end{tabular}
}
\end{table}

\subsubsection{Results on the FF Dataset} The face images in the FF Dataset have relatively poor visual qualities, simply because they are obtained from compressed video frames. Thus, the FF dataset is much more challenging to detect face image forgeries than the HFF dataset. To further prove the effectiveness of the proposed GRnet, similar experiments are conducted on the FF dataset for multiple and binary classifications. The experimental results are reported in Table VI and VII, respectively. From Table VI, we can observe that the shallow networks including Meso-4 and MesoInception-4 achieve only average accuracies of 51.78\% and 56.68\%, respectively. XceptionNet, which is a deep CNN model, also achieves only an average accuracy of 73.97\%, which is quite far away from satisfaction. The residuals based works including HighPass, MISLnet and AMTENnet achieve the average accuracies of 73.76\%, 74.03\% and 81.13\%, respectively.
For the binary classification, Table VII reports similar results to the multiple classification results reported in Table VI. Note that the main difference is that the average accuracy of the Capsule network is 0.23\% higher than that of GRnet for multiple classification, whereas it is 0.38\% lower than that of GRnet for binary classification.
It is well-known that compressed video frames launder the manipulation traces in the face image, which brings great challenges. By using capsule mechanism, Capsule network still maintains good modeling ability when dealing with low-quality data. Although the guided residuals are not applicable to low-quality data, we avoid the defects of the guided residuals as much as possible via the AFM, thus achieve the performance similar to that of the Capsule network.

\subsubsection{Results on the DFDC Dataset}
The DFDC dataset is made up of face images that captured from high-quality video frames. There are only two types of face images, namely real and fake. For face images in the DFDC dataset, it is a relatively easy task to make binary classification.
Table VIII reports the experimental results, which compares the ACC and AUC values among the proposed approach and the existing baseline works. From the results, the proposed GRnet still achieves the best accuracy of 99.09\% and the highest AUC value of 99.74\%.

\begin{table}[]
\centering
\footnotesize{TABLE VIII}
\caption*{\centering{\scriptsize{RESULTS FOR DIFFERENT FORENSICS MODELS}}}
\caption*{\centering{\scriptsize{ON THE DFDC DATASET}}}
\renewcommand{\arraystretch}{1.1}
\resizebox{60mm}{!}{
\begin{tabular}{|l|c|c|}
\hline
\multicolumn{1}{|c|}{Methods} & ACC              & AUC              \\ \hline
Meso-4\cite{MesoNet}                        & 70.47\%          & 71.62\%          \\ \hline
MesoInception-4\cite{MesoNet}               & 94.34\%          & 97.22\%          \\ \hline
HighPass\cite{binary_highpassfilter}                     & 93.37\%          & 93.35\%          \\ \hline
MISLnet\cite{constrainedCNN}                       & 96.62\%          & 96.65\%          \\ \hline
XceptionNet\cite{Xception}                   & 96.02\%          & 98.45\%          \\ \hline
Capsule\cite{CapsuleV2}                       & 98.04\%          & -          \\ \hline
AMTENnet\cite{AMTEN}                       & 96.64\%          & 96.66\%          \\ \hline
\hline

GRnet                  & \textbf{99.09\%} & \textbf{99.74\%} \\ \hline
\end{tabular}
}
\end{table}

\subsection{Generalization Ability}
To further validate the generalization ability of the model, we introduce the latest benchmark in Celeb-DF, which evaluates the generalization ability of existing methods on Celeb-DF dataset \cite{Celeb_DF}. The Celeb-DF dataset is a challenging dataset, which contains 5,639 hyper-realistic deepfake videos and 890 real face videos. We randomly select 30k real and 30k fake face frames from these videos as the test set. For the GRnet, the FF dataset with two quality levels is used to train it as a binary detector, which is referred to be GRnet-HQ and GRnet-LQ. They are then tested on the Celeb-DF dataset, respectively. Table IX compares the experimental results. We can observe that GRnet achieves better generalization ability than the existing works, especially GRnet-LQ achieves the highest AUC value up to 72.7\%.

\section{Conclusion}
Deepfake detection under complex Internet scenarios is a challenging problem in the community of image forensics due to diverse face image forgeries and low-quality visual content. In this work, we proposed a GRnet to detect the Deepfake face images under complex scenarios. Specifically, the MTE, which is an effective manipulation trace extractor, is introduced to obtain the guided residuals. Compared with the existing prediction based residuals, the guided residuals avoid the potential bias. In addition, we designed an AFM for feature fusion, which selectively emphasizes the relationship among feature channel maps and adaptively allocates the weights for two streams, further improving the detection performance. The extensive experimental results on four open datasets show that the proposed GRnet achieves much better accuracies and robustness than the state-of-the-art works, which is promising for face image forensics under complex scenarios.

\begin{table}[]
\centering
\footnotesize{TABLE IX}
\caption*{\centering{\scriptsize{THE AUC SCORES(\%) OF VARIOUS DETECTION METHODS}}}
\caption*{\centering{\scriptsize{ON CELEB-DF DATASET.}}}
\renewcommand{\arraystretch}{1.1}
\resizebox{80mm}{!}{
\begin{tabular}{|c|c|c|}
\hline
Methods                            & Training dataset        & Celeb-DF\cite{Celeb_DF}      \\ \hline
Two-stream\cite{zhoupeng}          & Private dataset         & 53.8\%                         \\ \hline
Meso-4\cite{MesoNet}               & Private dataset         & 54.8\%                         \\ \hline
MesoInception-4\cite{MesoNet}      & Private dataset         & 53.6\%                         \\ \hline
HeadPose\cite{head_poses}          & UADFV dataset           & 54.6\%                         \\ \hline
FWA\cite{SimulateArtifacts}        & UADFV dataset           & 56.9\%                         \\ \hline
VA-MLP\cite{visual_artifacts}      & Private dataset         & 55.0\%                         \\ \hline
VA-LogReg\cite{visual_artifacts}   & Private dataset         & 55.1\%                         \\ \hline
Xception-raw\cite{FaceForensics++} & FaceForensics++ dataset & 48.2\%                         \\ \hline
Xception-HQ\cite{FaceForensics++}  & FaceForensics++ dataset & 65.3\%                         \\ \hline
Xception-LQ\cite{FaceForensics++}  & FaceForensics++ dataset & 65.5\%                         \\ \hline
Multi-task\cite{Multi-task}        & FaceForensics dataset   & 54.3\%                         \\ \hline
Capsule\cite{CapsuleV2}              & Private dataset       & 57.5\%                         \\ \hline
\hline
GRnet-HQ                          & FaceForensics++ dataset & 54.6\%          \\ \hline
GRnet-LQ                          & FaceForensics++ dataset & \textbf{72.7\%} \\ \hline
\end{tabular}
}
\end{table}

\ifCLASSOPTIONcaptionsoff
  \newpage
\fi



\begin{thebibliography}{99}

\bibitem{CGR_tmm}
F. Peng, L. Yin, L. Zhang and M. Long. ``CGR-GAN: CG Facial Image Regeneration for Antiforensics Based on Generative Adversarial Network," \emph{IEEE Trans. Multimedia}, vol. 22, no. 10, pp. 2511-2525, Dec. 2019.

\bibitem{eye_blinking}
Y. Li, M. Chang and S. Lyu. ``In Ictu Oculi: Exposing AI Created Fake Videos by Detecting Eye Blinking," in \emph{Proc. IEEE Int. Workshop Inf. Forensics Security}, Dec. 2018, pp. 1-7.

\bibitem{head_poses}
X. Yang, Y. Li and S. Lyu. ``Exposing Deep Fakes Using Inconsistent Head Poses," in \emph{Proc. IEEE Int. Conf. Acoust., Speech, Signal Process.}, May. 2019, pp. 8261-8265.

\bibitem{visual_artifacts}
F. Matern, C. Riess and M. Stamminger. ``Exploiting Visual Artifacts to Expose Deepfakes and Face Manipulations," in \emph{Proc. IEEE Wint. Appl. Comput. Vis. Workshops}, Jan. 2019, pp. 83-92.

\bibitem{MesoNet}
D. Afchar, V. Nozick, J. Yamagishi, and I. Echizen. ``Mesonet: a compact facial video forgery detection network," in \emph{Proc. IEEE Int. Workshop Inf. Forensics Security}, Dec. 2018, pp. 1-7.

\bibitem{AppliedSciences}
L. M. Dang, S. I. Hassan, S. Im, J. Lee, S. Lee, and H. Moon. ``Deep learning based computer generated face identification using convolutional neural network," \emph{Appl. Sci.}, vol. 8, no. 12, pp. 2610-2628, Dec. 2018.

\bibitem{binary_highpassfilter}
H. Mo, B. Chen, and W. Luo. ``Fake faces identification via convolutional neural network," in \emph{Proc. 6th ACM Workshop Inf. Hid. Multimedia Security}, Jun. 2018, pp. 43-47.

\bibitem{AMTEN}
Z. Guo, G. Yang, J. Chen and X. Sun. ``Fake face detection via adaptive manipulation traces extraction network,"  \emph{Computer Vision and Image Understanding}, vol. 204, no. 12, pp. 103170, Jan. 2021.

\bibitem{forensics1_tmm}
X. Zhao, Y. Lin, and J. Heikkilä, ``Dynamic Texture Recognition Using Volume Local Binary Count Patterns With an Application to 2D Face Spoofing Detection," \emph{IEEE Trans. Multimedia}, vol. 20, no. 3, pp. 552-566, Mar. 2018.

\bibitem{forensics2_tmm}
X. Feng, I. J. Cox, and G. Doerr, ``Normalized Energy Density-Based Forensic Detection of Resampled Images," \emph{IEEE Trans. Multimedia}, vol. 14, no. 3, pp. 536-545, Jun. 2012.

\bibitem{pip1popescu}
A. C. Popescu, and H. Farid. ``Exposing digital forgeries by detecting traces of resampling," \emph{IEEE Trans. Signal Process.}, vol. 53, no. 2, pp. 758-767, Feb. 2005.

\bibitem{pip2qiu}
X. Qiu, H. Li, W. Luo, and J. Huang. ``A universal image forensic strategy based on steganalytic model," in \emph{Proc. 2th ACM Workshop Inf. Hid. Multimedia Security}, Jun. 2014, pp. 165-170.

\bibitem{pip3kirchner}
M. Kirchner. ``Fast and reliable resampling detection by spectral analysis of fixed linear predictor residue," in \emph{Proc. 10th ACM Workshop Multimedia Security}, New York, NY, USA, 2008, pp. 11-20.

\bibitem{Fingerprints}
N. Yu, L. Davis, and M. Fritz. ``Attributing Fake Images to GANs: Learning and Analyzing GAN Fingerprints," in \emph{Proc. ICCV.}, Oct. 2019.

\bibitem{DoGANs}
F. Marra, D. Gragnaniello, L. Verdoliva, and G. Poggi. ``Do GANs Leave Artificial Fingerprints," in \emph{Proc. IEEE Conf. Multimedia Inf. Process. Retrieval}, Mar. 2019, pp. 506-511.

\bibitem{guided_filter}
K. He, J. Sun and X. Tang, ``Guided Image Filtering," \emph{IEEE Trans. Pattern Anal. Mach. Intell.}, vol. 35, no. 6, pp. 1397-1409, Jun. 2013.

\bibitem{FaceForensics++}
A. R{\"o}ssler, D. Cozzolino, L. Verdoliva, C. Riess, J. Thies, and M. Nie{\ss}ner. ``FaceForensics++: Learning to Detect Manipulated Facial Images," in \emph{Proc. ICCV.}, Oct. 2019, pp. 1-11.

\bibitem{faceswap_tmm}
L. Liang, and X. Zhang. ``Adaptive Label Propagation for Facial Appearance Transfer," \emph{IEEE Trans. Multimedia}, vol. 21, no. 12, pp. 3068-3082, Dec. 2019.

\bibitem{glow}
D. P. Kingma, P. Dhariwal. ``Glow: Generative flow with invertible 1$\times$1 convolutions," in \emph{Proc. NIPS.}, Dec. 2018, pp. 10215--10224.

\bibitem{ResNet}
K. He, X. Zhang, S. Ren, and J. Sun. ``Deep Residual Learning for Image Recognition," in \emph{Proc. CVPR.}, Jun. 2016, pp. 770-778.

\bibitem{GANimation}
A. Pumarola, A. Agudo, A. M. Martinez, A. Sanfeliu, and F. Moreno-Noguer. ``GANimation: Anatomically-aware facial animation from a single image," in \emph{Proc. ECCV.}, Sep. 2018, pp. 818-833.

\bibitem{BEGAN}
D. Berthelot, T. Schumm, and L. Metz. (May. 2017). ``BEGAN: Boundary equilibrium generative adversarial networks." [Online]. Available: https://arxiv.org/abs/1703.10717

\bibitem{pggan}
T. Karras, T. Aila, S. Laine, and J. Lehtinen. ``Progressive Growing of GANs for Improved Quality, Stability, and Variation," in \emph{Proc. ICLR.}, Apr. 2018, pp. 1-26.

\bibitem{stylegan}
T. Karras, S. Laine, and T. Aila. ``A Style-Based Generator Architecture for Generative Adversarial Networks," in \emph{Proc. CVPR.}, Jun. 2019, pp. 4401-4410.

\bibitem{StarGAN}
Y. Choi, M. Choi, M. Kim, J. Ha, S. Kim, and J. Choo. ``Stargan: Unified generative adversarial networks for multi-domain image-to-image translation," in \emph{Proc. CVPR.}, Jun. 2018, pp. 8789-8797.

\bibitem{Face2Face}
J. Thies, M. Zollh\"{o}fer, M. Stamminger, C. Theobalt, and M. Nie{\ss}ner. ``Face2face: Real-time face capture and reenactment of RGB videos," in \emph{Proc. CVPR.}, Jun. 2016, pp. 2387-2395.

\bibitem{NeuralTextures}
J. Thies, M. Zollh{\"o}fer, and M. Nie{\ss}ner. ``Deferred neural rendering: Image synthesis using neural textures," \emph{ACM Transactions on Graphics}, 2019.

\bibitem{ARnet}
Y. Zhu, C. Chen, G. Yan, Y. Guo, and Y. Dong. ``AR-Net: Adaptive Attention and Residual Refinement Network for Copy-Move Forgery Detection," \emph{IEEE Trans. Ind. Informat.}, 2020.

\bibitem{SimulateArtifacts}
Y. Li, and S. Lyu. ``Exposing deepfake videos by detecting face warping artifacts," in \emph{Proc. CVPR. Workshops}, Jun. 2018, pp. 46-52.

\bibitem{ImbalanceData}
L. M. Dang, S. I. Hassan, S. Im, and H. Moon. ``Face image manipulation detection based on a convolutional neural network," \emph{Expert Syst. Appl.},  vol. 129, no. 1, pp. 156-168, Sep. 2019.

\bibitem{LandmarkSVM}
X. Yang, Y. Li, H. Qi, and S. Lyu. (Mar. 2019). ``Exposing GAN-synthesized Faces Using Landmark Locations.'' [Online]. Available: https://arxiv.org/abs/1904.00167

\bibitem{binary_lihaodong}
H. Li, B. Li, S. Tan, and J. Huang. (Aug. 2018). ``Detection of deep network generated images using disparities in color components.'' [Online]. Available: https://arxiv.org/abs/1808.07276

\bibitem{SRMfilter}
P. Zhou, X. Han, V. I. Morariu, and L. S. Davis. ``Learning Rich Features for Image Manipulation Detection," in \emph{Proc. CVPR.}, Jun. 2018, pp. 1053-1061.

\bibitem{constrainedCNN}
B. Bayar, and M. C. Stamm. ``Constrained convolutional neural networks: A new approach towards general purpose image manipulation detection," \emph{IEEE Trans. Inf. Forensics Security}, vol. 13, no. 11, pp. 2691-2706, Apr. 2018.

\bibitem{fusion_tmm}
X. Wang, L. Gao, P. Wang, X. Sun and X. Liu. ``Two-stream 3-d convnet fusion for action recognition in videos with arbitrary size and length," \emph{IEEE Trans. Multimedia}, vol. 20, no. 3, pp. 634-644, Mar. 2018.

\bibitem{sum_fusion}
K. Simonyan, and A. Zisserman. ``Two-Stream Convolutional Networks for Action Recognition in Videos," in \emph{Proc. NIPS.}, Dec. 2014, pp. 568-576.

\bibitem{fusion_strategy}
C. Feichtenhofer, A. Pinz, and A. Zisserman. ``Convolutional Two-Stream Network Fusion for Video Action Recognition," in \emph{Proc. CVPR.}, Jun. 2016, pp. 1933-1941.

\bibitem{att_first}
D. Bahdanau, K. Cho, and Y. Bengio. ``Neural Machine Translation by Jointly Learning to Align and Translate," in \emph{Proc. ICLR.}, May. 2015, pp. 1-15.

\bibitem{att_classification}
F. Wang, M. Jiang, C. Qian, S. Yang, C. Li, H. Zhang, X. Wang, and X. Tang. ``Residual Attention Network for Image Classification," in \emph{Proc. CVPR.}, Jul. 2017, pp. 3156-3164.

\bibitem{att_segmentation}
J. Fu, J. Liu, H. Tian, Y. Li, Y. Bao, Z. Fang, and H. Lu. ``Dual Attention Network for Scene Segmentation," in \emph{Proc. CVPR.}, Jun. 2019, pp. 3146-3154.

\bibitem{att_recognition}
J. Fu, H. Zheng, and T. Mei. ``Look Closer to See Better: Recurrent Attention Convolutional Neural Network for Fine-Grained Image Recognition," in \emph{Proc. CVPR.}, 2017, pp. 4438-4446.

\bibitem{non_local_att}
X. Wang, R. Girshick, A. Gupta, and K. He. ``Non-Local Neural Networks," in \emph{Proc. CVPR.}, Jun. 2018, pp. 7794-7803.

\bibitem{CrissCross_att}
Z. Huang, X. Wang, L. Huang, C. Huang, Y. Wei, and W. Liu. ``CCNet: Criss-Cross Attention for Semantic Segmentation," in \emph{Proc. ICCV.}, Oct. 2019, pp. 603-612.

\bibitem{ARnet}
Y. Zhu , C. Chen , G. Yan , Y. Guo and Y. Dong, ``AR-Net: Adaptive Attention and Residual Refinement Network for Copy-Move Forgery Detection," \emph{IEEE Trans. Ind. Informat.}, vol. 16, no. 10, pp. 6714-6723, Oct. 2020.

\bibitem{imagenet}
J. Deng, W. Dong, R. Socher, L. Li, Kai Li, and Li Fei-Fei. ``ImageNet: A large-scale hierarchical image database," in \emph{Proc. CVPR.}, 2009, pp. 248-255.

\bibitem{Forensics6fridrich2012}
J. Fridrich, and J. Kodovsky, ``Rich models for steganalysis of digital images," \emph{IEEE Trans. Inf. Forensics Security}, vol. 7, no. 3, pp. 868-882, Jun. 2012.

\bibitem{SPAM}
T. Pevny, P. Bas and J. Fridrich. ``Steganalysis by Subtractive Pixel Adjacency Matrix," \emph{IEEE Trans. Inf. Forensics Security}, vol. 5, no. 2, pp. 215-224, Jun. 2010.

\bibitem{MedResKang}
X. Kang, M. C. Stamm, A. Peng and K. J. R. Liu. ``Robust Median Filtering Forensics Using an Autoregressive Model," \emph{IEEE Trans. Inf. Forensics Security}, vol. 8, no. 9, pp. 1456-1468, Sep. 2013.

\bibitem{DFDC}
B. Dolhansky, R. Howes, B. Pflaum, N. Baram, and C. C. Ferrer. (Oct. 2019) ``The Deepfake Detection Challenge (DFDC) Preview Dataset." [Online]. Available: https://arxiv.org/abs/1910.08854

\bibitem{celeba}
Z. Liu, P. Luo, X. Wang, and X. Tang. ``Deep learning face attributes in the wild," in \emph{Proc. ICCV.}, Dec. 2015, pp. 3730-3738.

\bibitem{Faceforensics}
A. R{\"o}ssler, D. Cozzolino, L. Verdoliva, C. Riess, J. Thies, and M. Nie{\ss}ner. (Mar. 2018). ``FaceForensics: A large-scale video dataset for forgery detection in human faces." [Online]. Available: https://arxiv.org/abs/1803.09179

\bibitem{Xception}
F. Chollet. ``Xception: Deep Learning With Depthwise Separable Convolutions," in \emph{Proc. CVPR.}, Jul. 2017, pp. 1251-1258.

\bibitem{CapsuleV2}
H. H. Nguyen, J. Yamagishi, and I. Echizen. (Oct. 2019). ``Use of a capsule network to detect fake images and videos." [Online]. Available: https://arxiv.org/abs/1910.12467

\bibitem{visualization}
J. T. Springenberg, A. Dosovitskiy, T. Brox, and M. Riedmiller. (Dec. 2014). ``Striving for simplicity: The all convolutional net." [Online]. Available: https://arxiv.org/abs/1412.6806

\bibitem{Celeb_DF}
Y. Li, X. Yang, P. Sun, H. Qi, and S. Lyu. ``Celeb-DF: A Large-scale Challenging Dataset for DeepFake Forensics," in \emph{Proc. CVPR.}, 2020, pp. 3207-3216.

\bibitem{zhoupeng}
P. Zhou, X. Han, V. I. Morariu, and L. S. Davis. ``Two-stream neural networks for tampered face detection," in \emph{Proc. CVPR. Workshops}, Jul. 2017, pp. 1831-1839.

\bibitem{Multi-task}
H. H. Nguyen, F. Fang, J. Yamagishi, and I. Echizen. ``Multi-task learning for detecting and segmenting manipulated facial images and videos," in \emph{Proc. BTAS.}, 2019.


\end{thebibliography}
\end{document}